\pdfoutput=1

\documentclass[pdflatex,sn-nature]{sn-jnl}% Style for submissions to Nature Portfolio journals
\ifx\pdfminorversion\undefined\else\pdfminorversion=6\fi

\usepackage{graphicx}%
\usepackage{amsmath,amssymb,amsfonts}%
\usepackage{bm}%
\usepackage{placeins}%
\usepackage{booktabs}%
\usepackage{array}%
\usepackage{algorithm}%
\usepackage{algorithmic}%
\usepackage{color}%
\usepackage{multirow}%
\usepackage{siunitx}%
\usepackage{tikz}%
\usetikzlibrary{arrows,positioning,fit}

\graphicspath{{./}{figs/}}
\hypersetup{hypertexnames=false}

\theoremstyle{thmstyleone}%
\newtheorem{theorem}{Theorem}

\newcommand{\best}[1]{\bm{#1}}

\newcommand{\resulttablefont}{\small}

\newcommand{\unnumlead}[1]{\par\medskip\noindent\textbf{#1}\par\smallskip}
\newcommand{\methodlead}[1]{\unnumlead{#1}}

\begin{document}

\title[Photonic quantum neural fields for scientific machine learning]{
Photonic Quantum Neural Fields for Physics-Informed Scientific Machine Learning
}

\author[1]{\fnm{Jiale} \sur{Linghu}}
\author[1]{\fnm{Hao} \sur{Dong}}
\author*[2]{\fnm{Yangshuai} \sur{Wang}}

\affil[1]{\orgdiv{School of Mathematics and Statistics}, \orgname{Xidian University}, \orgaddress{\city{Xi'an}, \state{Shaanxi}, \postcode{710071}, \country{China}}}
\affil*[2]{\orgdiv{Department of Mathematics, Faculty of Science}, \orgname{National University of Singapore}, \orgaddress{\city{Singapore}, \postcode{119077}, \country{Singapore}}}

\abstract{
Photonic quantum machine learning offers a route to trainable physical representations built from phase, interference and measurement, but its role in scientific machine learning remains largely unexplored. Physics-informed neural fields provide a natural testbed, because differential equations require trial spaces that preserve phase, frequency and derivative structure. Here we introduce a photonic quantum neural field in which coordinates become trainable optical phases, are mixed by multi-photon Fock-space interference and are decoded from photon-number measurements. The photonic circuit is optimized as the neural-field representation itself, rather than used as a fixed feature map or hardware accelerator. Across seven elliptic, wave, nonlinear dispersive and inverse PDE benchmarks, we observe a phase-complexity transition: classical coordinate and Fourier-feature networks suffice in smooth regimes, whereas the photonic field is most accurate when residual derivatives amplify phase mismatch. In the hardest regimes it gives the lowest errors, by up to order-of-magnitude margins, using about one quarter of the trainable parameters of classical baselines. Frozen and shuffled circuit controls show that the gain depends on learned photonic interference, and noise stress tests show that Fock-probability readout is more stable than a qubit-style quantum baseline under compound perturbations. These results identify photonic quantum measurement as a representation-learning principle for scientific machine learning.
}


\maketitle

% ============================================================
A central question in machine learning is how to build representations that carry the structure of the systems they model. In scientific machine learning, this question is not only statistical but physical: useful representations must preserve symmetries, phases, constraints and derivative relations. Photonic quantum machine learning offers a route to such representations in which the trainable variables are physical: phase, interference, propagation and measurement \cite{obrien2009photonic,peruzzo2014vqe_photonic,wang2020integrated_photonic,pelucchi2022integrated_photonics,madsen2022photonic_advantage}. This prospect is becoming increasingly concrete as integrated photonics, scalable interferometric meshes and photon-number readout mature into programmable computational platforms. Integrated photonic neural networks are also beginning to support training as a physical process. A recent on-chip backpropagation demonstration showed that a photonic neural network can perform all linear and nonlinear computations on a single chip and compensate for fabrication-induced device variations during training \cite{ashtiani2026integrated_pnn_bp}. These developments give photonics a distinctive position in machine learning. It is not only a route to faster inference hardware, but a trainable physical medium whose native variables include phase, interference and modal mixing.

The representation problem is especially acute in scientific machine learning, where models are asked to solve equations, infer parameters and extrapolate physical fields from sparse data. Many of these tasks are governed by partial differential equations (PDEs), whose solutions are organized by frequency, phase, boundary constraints and derivative relations. Physics-informed neural networks (PINNs) exploit this structure by fitting differentiable neural fields through residual, boundary, initial and data losses \cite{raissi2019pinn,karniadakis2021piml}. Their appeal is that sparse observations, physical constraints and unknown parameters can be combined in one optimization problem. Their limitation is equally structural: the solver can only enforce the physics through the trial space it is given. For oscillatory, wave, dispersive and multiscale equations, that trial space must preserve phase and derivative structure, not only function values.

Existing quantum-learning work has established measured quantum states as kernels, feature maps and classifiers \cite{biamonte2017qml,havlicek2019quantum,schuld2019feature,perezsalinas2020data}. Scientific PDE solvers, by contrast, still rely mainly on classical coordinate networks or prescribed spectral embeddings. The missing capability is more specific than quantum feature generation in general. Multi-photon interference and Fock-space measurement have not been established as the solution representation on which a PDE residual itself is minimized.

The need for such a representation follows from the failure mode of physics-informed learning. In oscillatory regimes, differential operators convert small phase mismatches into residual-scale errors, with amplification increasing with frequency and derivative order. Standard coordinate networks are poorly matched to this regime because multilayer perceptrons tend to fit low-frequency components before high-frequency components, a behavior often described as spectral bias \cite{rahaman2019spectral}. In PINNs, the differential operator amplifies this bias because high-frequency and multiscale components often lie in poorly conditioned directions of the induced optimization dynamics \cite{wang2022ntk,wang2021eigenvector,krishnapriyan2021failure}. The obstacle is therefore not only capacity, but whether the representation places phase, frequency and derivative information in directions that the residual can learn.

Fourier-feature and random-feature methods address part of the mismatch by embedding coordinates into trigonometric feature spaces before learning the solution \cite{tancik2020fourier,rahimi2007rff,wang2021eigenvector}. These embeddings make high-frequency structure visible to a neural network. Yet in PDE residual minimization, the relevant frequencies, phase shifts, nonlinear interactions and bandwidth are often determined by the equation, boundary data and unknown physical parameters. A fixed, random or hand-selected frequency dictionary therefore leaves a central part of the representation outside the physics-informed optimization.

Parameterized quantum circuits offer a route to making this representation trainable. Classical inputs can be embedded into high-dimensional Hilbert spaces, where measurements define nonlinear features or kernels \cite{biamonte2017qml,schuld2019feature,havlicek2019quantum}. Data reuploading further increases expressive power by interleaving repeated input encoding with trainable circuit operations \cite{perezsalinas2020data}. In the photonic setting, this feature-space view becomes physically explicit. A coordinate can be encoded as an optical phase, producing factors such as $e^{iz}$; trainable interferometers coherently mix these phase factors across modes; multi-photon Fock states create combinatorial path interference; and Fock-space probability measurements convert complex amplitudes into real nonlinear spectral moments.

The resulting photonic map is Fourier-like, but not a static Fourier basis. Its accessible spectrum depends on circuit topology, photon number, input state, trainable phases, measurement choice and reuploading depth, while the measured features remain bounded and physically normalized. Modern photonic software ecosystems make these circuits increasingly accessible as differentiable components for machine-learning workflows \cite{killoran2019strawberry,heurtel2023perceval,notton2026merlin}. This combination of spectral structure, trainability and measurement-defined nonlinearity is precisely what phase-sensitive residual learning requires.

Here we replace the coordinate-network trial space in a PINN by a photonic spectral generator, yielding a physics-informed hybrid photonic quantum neural network (PI-HPQNN) for PDEs. The architecture keeps the PINN principle of residual minimization, but changes the object being optimized. Physical coordinates are lifted by a classical network and converted into wrapped optical phases. These phases drive a multi-mode Fock-space circuit, whose photon-number probabilities are decoded into the PDE field. The full composition is differentiable, so interference, measurement and data reuploading are optimized by the same residual, boundary, data and parameter-identification losses as the classical baselines. The photonic circuit is therefore not an auxiliary feature extractor. It is the measured trial representation on which the residual acts.

We test this representation across elliptic, wave, nonlinear dispersive and inverse benchmarks, where its advantage emerges selectively rather than uniformly. PI-HPQNN is strongest when residual accuracy depends on phase-consistent derivatives rather than smooth interpolation alone, including high-frequency Poisson and Helmholtz problems, forced oscillatory dynamics, complex nonlinear Schr\"odinger fields and sparse or noisy inverse recovery. Structure ablations show that accuracy is controlled by which measured spectral moments the circuit makes available, while frozen and shuffled circuit controls show that the gain depends on learned photonic interference. An evaluation-time noise stress test further shows that normalized Fock-probability features are more stable than qubit expectation features under compound transmission, visibility and shot perturbations. Together, these results identify photonic quantum measurement not merely as a readout or hardware primitive, but as a trainable representation-learning principle for scientific fields organized by phase.

% ============================================================

% ============================================================

% ============================================================
\section{Results}
% ============================================================

\unnumlead{Testing a measured photonic representation for PDE residual learning}

\begin{figure}[!htbp]
\centering
\includegraphics[width=\textwidth]{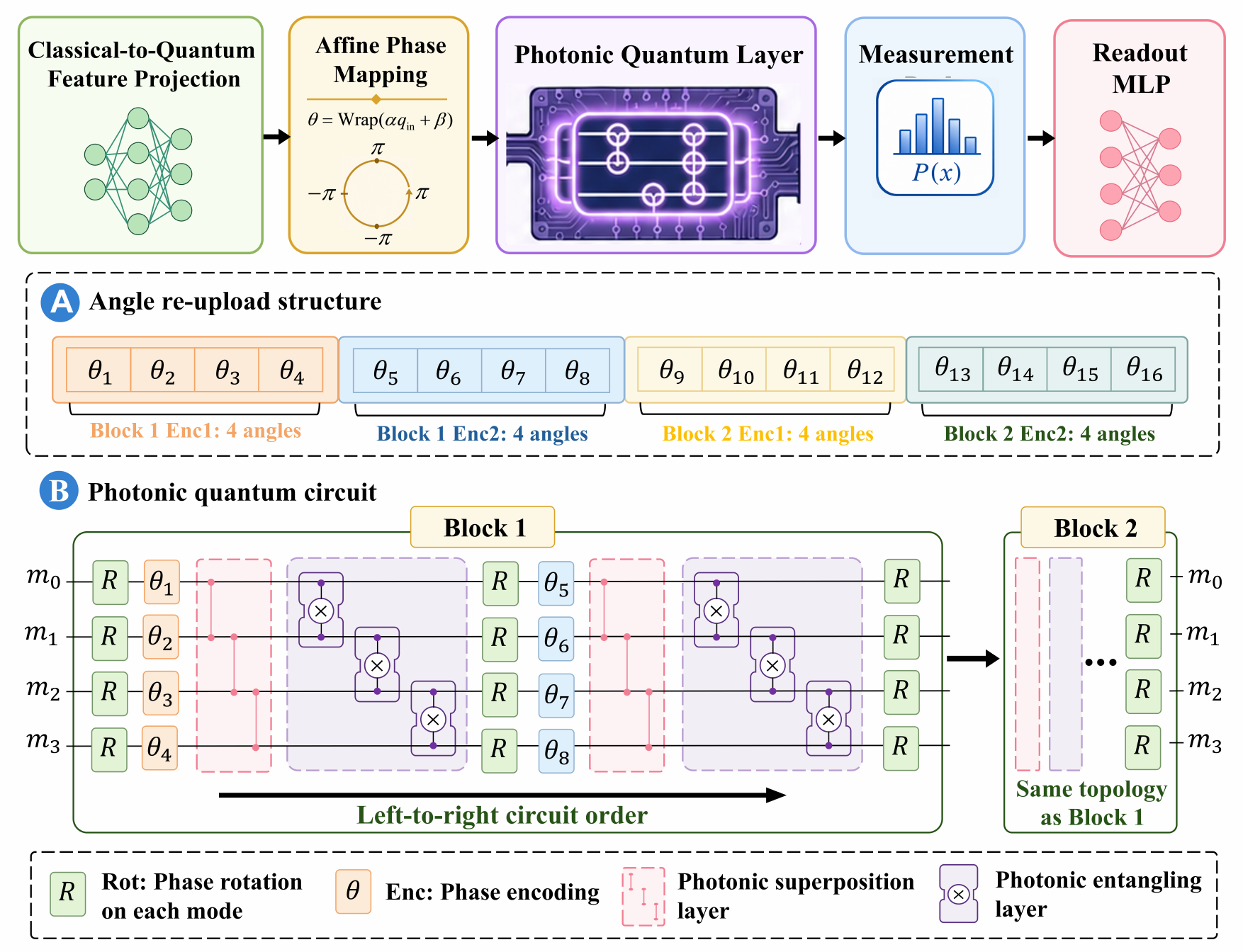}
\caption{Physics-informed photonic quantum neural field. Physical coordinates are mapped to optical phase variables, processed by a two-block MerLin-based photonic quantum layer with repeated phase encoding and multi-mode interference, measured as Fock probabilities and decoded into the PDE solution. The physics-informed loss is evaluated on the final differentiable field, allowing the phase encoder, photonic circuit and classical readout to be trained jointly.}
\label{fig:results_framework}
\end{figure}

We first tested whether a photonic quantum circuit can serve as the trainable solution representation in a physics-informed PDE solver, rather than as a fixed feature map or a hardware accelerator. PI-HPQNN implements this test by replacing the coordinate-network trial space with a measured photonic neural field (Fig.~\ref{fig:results_framework}). Physical coordinates are lifted into trainable optical phases, mixed by fixed-photon-number interference and decoded from Fock-space probabilities before the PDE residual is evaluated.

This setup changes the object being optimized. In a conventional physics-informed neural network, the residual trains a prechosen coordinate representation. In PI-HPQNN, the same residual, boundary and data losses train the phase encoder, photonic interferometer and classical readout jointly. The photonic circuit therefore supplies part of the differentiable field on which the residual acts.

The benchmark comparisons isolate this coordinate-to-field representation. We trained standard coordinate networks, fixed Fourier-feature networks, a hybrid quantum neural network baseline and PI-HPQNN under matched residual losses and protocols. The central question was not whether the photonic model wins every PDE task, but whether the learned measurement representation becomes useful when the residual amplifies phase error.

\unnumlead{Accuracy reverses in phase-complex PDE regimes}

The forward benchmarks were organized as phase-complexity ladders (Table~\ref{tab:forward_problem_summary}). In each equation family, one control parameter increases oscillatory content and residual sensitivity to phase mismatch: wavenumber \(\lambda\) in Poisson and Helmholtz, phase rate \(b\) in the wave equation, forcing wavenumber \(k\) in Sine--Gordon and plane-wave index \(n\) in NLS. This design produced a consistent reversal rather than a uniform win. At low complexity, classical coordinate and Fourier-feature baselines already resolve the solution at the \(10^{-3}\) level or below, and PI-HPQNN offers no advantage. The best model is classical by factors of \(1.4\)--\(4.3\) at \(\lambda=1\) (Poisson), \(b\in\{1,2\}\) (wave), \(\lambda=2\) (Helmholtz), \(k\in\{6,7\}\) (Sine--Gordon) and \(n=2\) (NLS).

Beyond a problem-dependent threshold, the ordering inverts. PI-HPQNN becomes the most accurate model at \(\lambda\ge 5\) for Poisson and Helmholtz, \(b\ge 3\) for the wave equation, \(k\ge 8\) for Sine--Gordon and \(n\ge 4\) for NLS. In these regimes, it beats the strongest classical baseline by factors ranging from \(1.5\times\) in the wave benchmark to \(12\times\) in Helmholtz and NLS amplitude recovery, while using roughly one quarter of the trainable parameters of the classical baselines. The repeated pattern is the main result of Table~\ref{tab:forward_problem_summary}: the photonic representation is not generically superior, but becomes superior when phase mismatch is derivative-amplified.

This crossover identifies when the photonic trial space matters. Below the threshold, differential operators do not strongly amplify representation error, and static coordinate or fixed-Fourier embeddings are sufficient. Above the threshold, residual derivatives convert small phase mismatches into residual-scale errors. The measured photonic representation then supplies trainable spectral moments that classical baselines do not expose. This gives a practical design rule: use the photonic trial space when the dominant wavenumber or time-dependent phase rate lies in the derivative-amplified regime; below that scale, classical baselines remain preferable.

\begin{table}[!htbp]
\centering
\fontsize{4.5}{5.2}\selectfont
\caption{Forward-problem phase-complexity benchmark. Entries are relative \(L^2\) errors reported as mean \(\pm\) sample standard deviation. For NLS, the coupled-field error and the amplitude error are reported as separate metric rows. Boldface marks the lowest mean error in each row.}
\label{tab:forward_problem_summary}
\setlength{\tabcolsep}{0pt}
\renewcommand{\arraystretch}{1.18}
\begin{tabular}{lllcccc}
\toprule
Equation & Parameter value & Metric
& \multicolumn{1}{c}{PINN}
& \multicolumn{1}{c}{FPINN}
& \multicolumn{1}{c}{HQNN}
& \multicolumn{1}{c}{PI-HPQNN} \\
\midrule
\multirow{3}{*}{Poisson}
& \(\lambda=1\) & Rel. \(L^2\) & \((9.59 \pm 3.56){\times}10^{-5}\) & \(\best{(9.40 \pm 4.38){\times}10^{-5}}\) & \((1.45 \pm 0.13){\times}10^{-4}\) & \((3.28 \pm 1.82){\times}10^{-4}\) \\
& \(\lambda=5\) & Rel. \(L^2\) & \((1.66 \pm 0.26){\times}10^{0}\) & \((4.61 \pm 1.30){\times}10^{-2}\) & \((3.18 \pm 4.65){\times}10^{-1}\) & \(\best{(6.11 \pm 1.73){\times}10^{-3}}\) \\
& \(\lambda=6\) & Rel. \(L^2\) & \((2.81 \pm 0.83){\times}10^{0}\) & \((9.00 \pm 4.65){\times}10^{-2}\) & \((1.76 \pm 0.92){\times}10^{-1}\) & \(\best{(3.05 \pm 1.28){\times}10^{-2}}\) \\
\addlinespace[0.25em]
\multirow{3}{*}{Wave}
& \(b=1\) & Rel. \(L^2\) & \(\best{(2.23 \pm 0.48){\times}10^{-3}}\) & \((3.36 \pm 0.65){\times}10^{-3}\) & \((9.24 \pm 2.63){\times}10^{-3}\) & \((8.75 \pm 0.65){\times}10^{-3}\) \\
& \(b=2\) & Rel. \(L^2\) & \(\best{(1.33 \pm 0.18){\times}10^{-2}}\) & \((1.87 \pm 0.15){\times}10^{-2}\) & \((3.05 \pm 0.47){\times}10^{-2}\) & \((2.38 \pm 0.63){\times}10^{-2}\) \\
& \(b=3\) & Rel. \(L^2\) & \((5.06 \pm 3.14){\times}10^{-1}\) & \((1.09 \pm 0.43){\times}10^{-1}\) & \((2.75 \pm 1.51){\times}10^{-1}\) & \(\best{(7.26 \pm 1.46){\times}10^{-2}}\) \\
\addlinespace[0.25em]
\multirow{3}{*}{Helmholtz}
& \(\lambda=2\) & Rel. \(L^2\) & \((1.99 \pm 0.44){\times}10^{-3}\) & \((2.66 \pm 1.48){\times}10^{-3}\) & \(\best{(1.47 \pm 0.36){\times}10^{-3}}\) & \((6.32 \pm 1.15){\times}10^{-3}\) \\
& \(\lambda=3\) & Rel. \(L^2\) & \((3.69 \pm 4.24){\times}10^{-2}\) & \((2.49 \pm 1.58){\times}10^{-2}\) & \((4.65 \pm 6.25){\times}10^{-2}\) & \(\best{(9.51 \pm 4.15){\times}10^{-3}}\) \\
& \(\lambda=4\) & Rel. \(L^2\) & \((2.24 \pm 0.89){\times}10^{0}\) & \((3.18 \pm 0.77){\times}10^{-1}\) & \((4.39 \pm 0.68){\times}10^{-1}\) & \(\best{(2.72 \pm 1.34){\times}10^{-2}}\) \\
\addlinespace[0.25em]
\multirow{3}{*}{Sine--Gordon}
& \(k=6\) & Rel. \(L^2\) & \((9.75 \pm 8.32){\times}10^{-1}\) & \(\best{(1.81 \pm 0.46){\times}10^{-2}}\) & \((5.47 \pm 3.75){\times}10^{-2}\) & \((5.69 \pm 2.83){\times}10^{-2}\) \\
& \(k=7\) & Rel. \(L^2\) & \((1.76 \pm 0.53){\times}10^{0}\) & \(\best{(4.24 \pm 0.54){\times}10^{-2}}\) & \((1.66 \pm 1.03){\times}10^{-1}\) & \((6.02 \pm 2.65){\times}10^{-2}\) \\
& \(k=8\) & Rel. \(L^2\) & \((2.13 \pm 1.05){\times}10^{0}\) & \((8.33 \pm 1.10){\times}10^{-1}\) & \((6.51 \pm 8.60){\times}10^{-1}\) & \(\best{(8.71 \pm 3.72){\times}10^{-2}}\) \\
\addlinespace[0.25em]
\multirow{6}{*}{NLS}
& \multirow{2}{*}{\(n=2\)} & \((u,v)\) & \((3.02 \pm 0.49){\times}10^{-3}\) & \(\best{(1.78 \pm 0.16){\times}10^{-3}}\) & \((2.93 \pm 0.77){\times}10^{-3}\) & \((3.76 \pm 1.07){\times}10^{-3}\) \\
& & \(|\psi|\) & \((1.88 \pm 0.38){\times}10^{-3}\) & \(\best{(1.26 \pm 0.08){\times}10^{-3}}\) & \((1.86 \pm 0.33){\times}10^{-3}\) & \((2.23 \pm 0.51){\times}10^{-3}\) \\
\addlinespace[0.15em]
& \multirow{2}{*}{\(n=4\)} & \((u,v)\) & \((3.40 \pm 1.29){\times}10^{-2}\) & \((4.80 \pm 4.90){\times}10^{-1}\) & \((2.67 \pm 0.83){\times}10^{-2}\) & \(\best{(2.38 \pm 0.66){\times}10^{-2}}\) \\
& & \(|\psi|\) & \((1.72 \pm 0.49){\times}10^{-2}\) & \((3.13 \pm 3.07){\times}10^{-1}\) & \((1.76 \pm 0.79){\times}10^{-2}\) & \(\best{(1.35 \pm 0.31){\times}10^{-2}}\) \\
\addlinespace[0.15em]
& \multirow{2}{*}{\(n=6\)} & \((u,v)\) & \((1.05 \pm 0.09){\times}10^{0}\) & \((9.92 \pm 0.08){\times}10^{-1}\) & \((5.19 \pm 0.86){\times}10^{-1}\) & \(\best{(6.67 \pm 3.86){\times}10^{-2}}\) \\
& & \(|\psi|\) & \((7.86 \pm 2.93){\times}10^{-1}\) & \((8.97 \pm 0.18){\times}10^{-1}\) & \((3.83 \pm 0.71){\times}10^{-1}\) & \(\best{(3.12 \pm 2.04){\times}10^{-2}}\) \\
\bottomrule
\end{tabular}
\end{table}

\begin{figure}[!htbp]
\centering
\includegraphics[width=\textwidth]{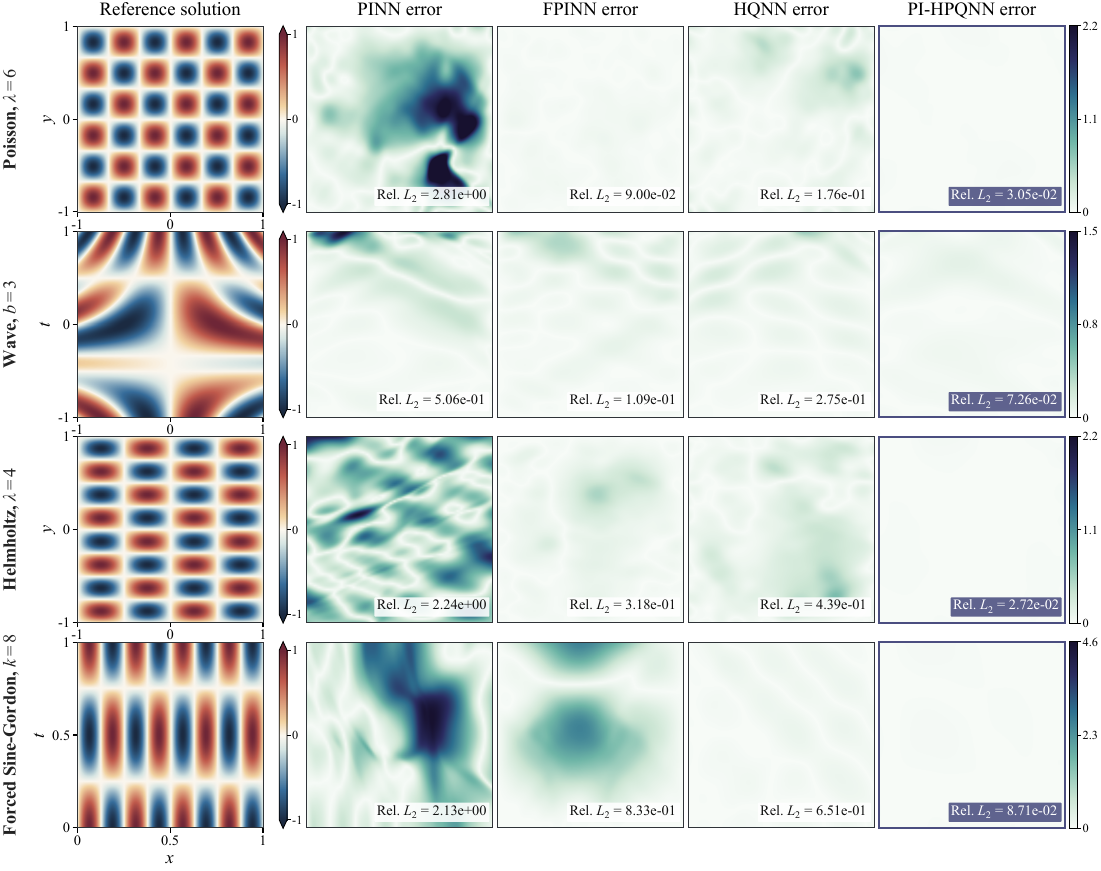}
\caption{Error-field comparison in representative phase-complex forward benchmarks. The rows show the reference solution and absolute-error maps for Poisson at \(\lambda=6\), the wave equation at \(b=3\), Helmholtz at \(\lambda=4\), and forced Sine--Gordon at \(k=8\). Across elliptic, wave and nonlinear oscillatory settings, PI-HPQNN suppresses coherent phase-aligned error structures, showing that the photonic representation improves the geometry of the learned field rather than only the aggregate error value.}
\label{fig:forward_solution_composite}
\end{figure}

The forward reconstructions reveal the geometric form of this advantage. Figure~\ref{fig:forward_solution_composite} compares the spatial error fields in four representative phase-complex settings: high-frequency Poisson, rapidly varying wave dynamics, high-wavenumber Helmholtz and forced Sine--Gordon. The dominant difference is not only the magnitude of the error, but the structure of the remaining error field. Classical coordinate networks leave coherent bands or islands of error, a signature that the learned solution has slipped in phase relative to the target field. Fixed Fourier features can reduce this mismatch when the prescribed dictionary happens to match the solution spectrum, but structured error remains when the residual requires a richer phase-consistent trial field. PI-HPQNN suppresses these coherent error modes across the four equation classes. The figure therefore visualizes the central mechanism of the method: multi-photon interference and Fock-space measurement generate spectral features that keep the field and its derivatives aligned under the PDE residual, turning photonic phase processing into a compact solution representation rather than a larger classical parametrization.

\unnumlead{Photonic features sharpen parameter recovery from sparse constraints}

The inverse benchmarks ask a more targeted question: can the measured photonic representation help sparse observations identify the governing parameter, even when pointwise field interpolation is not always best? Table~\ref{tab:inverse_summary} combines the inverse Burgers and Euler tests. In Burgers, classical baselines give the smallest field errors, but PI-HPQNN gives the smallest viscosity error in both noiseless and \(5\%\)-noise settings. The useful signal is therefore parameter recovery rather than blanket state accuracy. The Euler benchmark is more tightly coupled, requiring simultaneous recovery of density, velocity, pressure and the coefficient \(k\). In this setting PI-HPQNN gives the closest coefficient estimate and the smallest state errors in both data regimes. Together, the inverse tests show that measured photonic features can couple sparse field observations to latent physical parameters, which is the inverse-problem analogue of the phase-complex forward transition.

\begin{table}[!htbp]
\centering
\fontsize{4.5}{5.2}\selectfont
\caption{Inverse-problem benchmark summary. For Burgers, the coefficient column reports the relative viscosity error and the state column reports the solution relative \(L^2\) error. For Euler, the coefficient column reports the estimated \(k\), and the state columns report relative errors for \(\rho\), \(u\) and \(p\). Boldface marks the lowest error or closest coefficient estimate in each case.}
\label{tab:inverse_summary}
\setlength{\tabcolsep}{1.2pt}
\renewcommand{\arraystretch}{1.08}
\begin{tabular}{lllcccc}
\toprule
Benchmark & Case & Model & Coefficient metric & State 1 & State 2 & State 3 \\
\midrule
\multirow{8}{*}{Burgers}
& \multirow{4}{*}{No noise}
& PINN & \((2.60 \pm 1.96){\times}10^{-2}\) & \((6.27 \pm 2.19){\times}10^{-3}\) & -- & -- \\
& & FPINN & \((2.76 \pm 0.83){\times}10^{-2}\) & \(\best{(4.79 \pm 0.68){\times}10^{-3}}\) & -- & -- \\
& & HQNN & \((5.00 \pm 2.37){\times}10^{-2}\) & \((1.05 \pm 0.23){\times}10^{-2}\) & -- & -- \\
& & PI-HPQNN & \(\best{(6.04 \pm 4.11){\times}10^{-3}}\) & \((1.70 \pm 0.17){\times}10^{-2}\) & -- & -- \\
\addlinespace[0.25em]
& \multirow{4}{*}{\(5\%\) noise}
& PINN & \((3.30 \pm 1.60){\times}10^{-2}\) & \(\best{(9.94 \pm 1.65){\times}10^{-3}}\) & -- & -- \\
& & FPINN & \((2.78 \pm 3.54){\times}10^{-2}\) & \((1.25 \pm 0.18){\times}10^{-2}\) & -- & -- \\
& & HQNN & \((4.38 \pm 3.16){\times}10^{-2}\) & \((1.59 \pm 0.41){\times}10^{-2}\) & -- & -- \\
& & PI-HPQNN & \(\best{(1.78 \pm 0.85){\times}10^{-2}}\) & \((1.86 \pm 0.35){\times}10^{-2}\) & -- & -- \\
\addlinespace[0.35em]
\multirow{8}{*}{Euler}
& \multirow{4}{*}{No noise}
& PINN & \(0.9952 \pm 0.0130\) & \((3.15 \pm 2.55){\times}10^{-2}\) & \((6.43 \pm 3.26){\times}10^{-4}\) & \((5.53 \pm 3.22){\times}10^{-4}\) \\
& & FPINN & \(0.9984 \pm 0.0055\) & \((6.38 \pm 1.71){\times}10^{-3}\) & \((1.99 \pm 1.68){\times}10^{-3}\) & \((3.29 \pm 3.58){\times}10^{-3}\) \\
& & HQNN & \(0.9973 \pm 0.0025\) & \((3.69 \pm 1.01){\times}10^{-2}\) & \((9.11 \pm 3.53){\times}10^{-4}\) & \((1.24 \pm 0.39){\times}10^{-3}\) \\
& & PI-HPQNN & \(\best{0.9996 \pm 0.0010}\) & \(\best{(1.35 \pm 0.47){\times}10^{-3}}\) & \(\best{(1.11 \pm 0.38){\times}10^{-4}}\) & \(\best{(1.19 \pm 0.45){\times}10^{-4}}\) \\
\addlinespace[0.25em]
& \multirow{4}{*}{\(5\%\) noise}
& PINN & \(1.0212 \pm 0.0183\) & \((2.60 \pm 2.51){\times}10^{-2}\) & \((4.91 \pm 1.49){\times}10^{-4}\) & \((5.15 \pm 1.55){\times}10^{-4}\) \\
& & FPINN & \(0.9985 \pm 0.0183\) & \((1.45 \pm 0.53){\times}10^{-2}\) & \((1.41 \pm 0.42){\times}10^{-3}\) & \((1.77 \pm 0.70){\times}10^{-3}\) \\
& & HQNN & \(0.9849 \pm 0.0189\) & \((2.52 \pm 2.30){\times}10^{-2}\) & \((7.02 \pm 3.95){\times}10^{-4}\) & \((6.96 \pm 4.48){\times}10^{-4}\) \\
& & PI-HPQNN & \(\best{1.0013 \pm 0.0155}\) & \(\best{(8.56 \pm 8.32){\times}10^{-3}}\) & \(\best{(2.84 \pm 1.49){\times}10^{-4}}\) & \(\best{(2.45 \pm 0.68){\times}10^{-4}}\) \\
\bottomrule
\end{tabular}
\end{table}

\unnumlead{A compact photonic representation reduces trainable dimension}

\begin{figure}[!htbp]
\centering
\includegraphics[width=\textwidth]{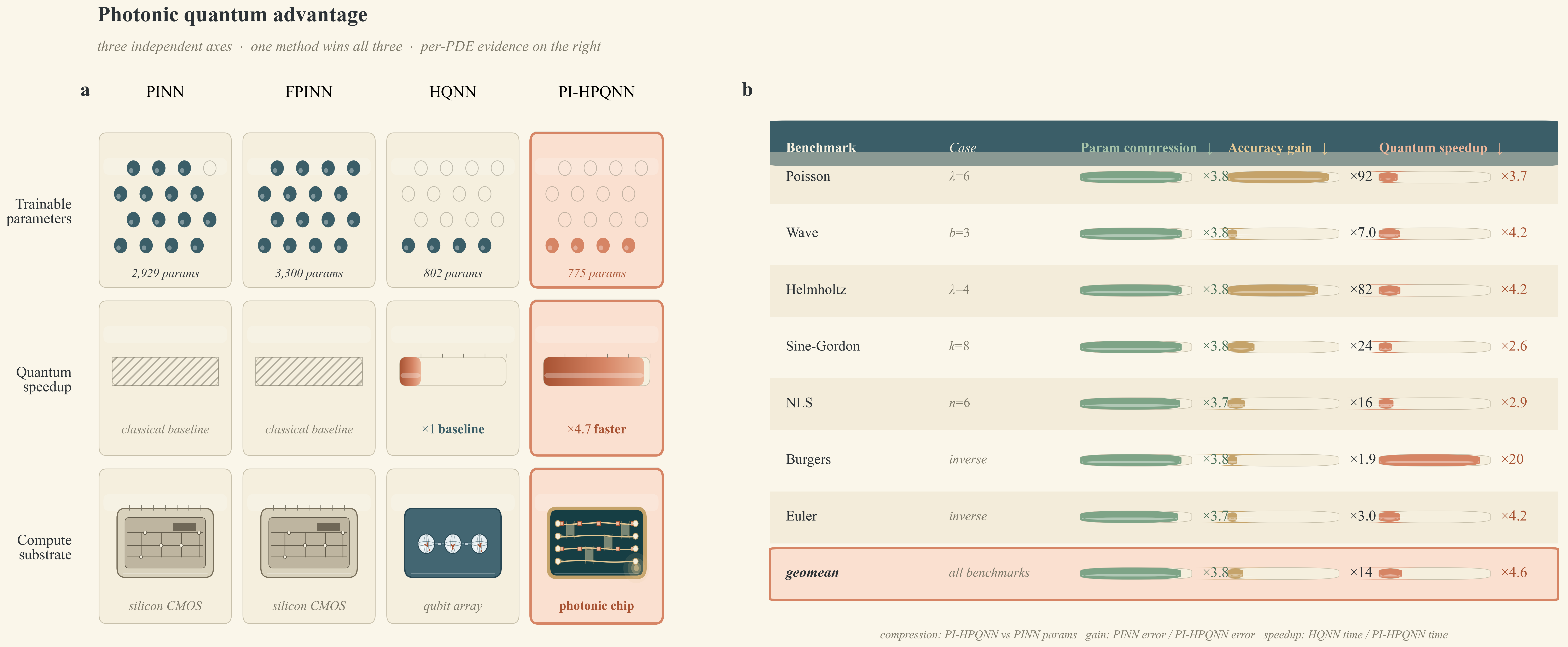}
\caption{Parameter and training-time profile under differentiable classical simulation. PI-HPQNN uses a compact parameter budget across the benchmark suite, whereas the reported wall-clock times reflect classical simulation of the photonic circuit rather than native photonic hardware execution.}
\label{fig:efficiency_summary}
\end{figure}

The phase-complex gains are not obtained by increasing the trainable dimension. Figure~\ref{fig:efficiency_summary} summarizes the parameter and training-time profile across the benchmark suite. PI-HPQNN uses roughly one quarter of the trainable parameters used by the classical coordinate and Fourier-feature baselines, while still giving the strongest accuracy in the phase-complex cases. This supports the interpretation that the photonic layer changes the representation, not merely the model size. The wall-clock times in this figure are differentiable classical simulation times, not native photonic hardware timings. They therefore document the software cost of simulating multi-photon interference, while keeping the parameter-efficiency claim separate from any hardware-throughput claim.

\unnumlead{The learned measurement representation exposes phase-aligned moments}

\begin{figure}[!htbp]
\centering
\includegraphics[width=\textwidth]{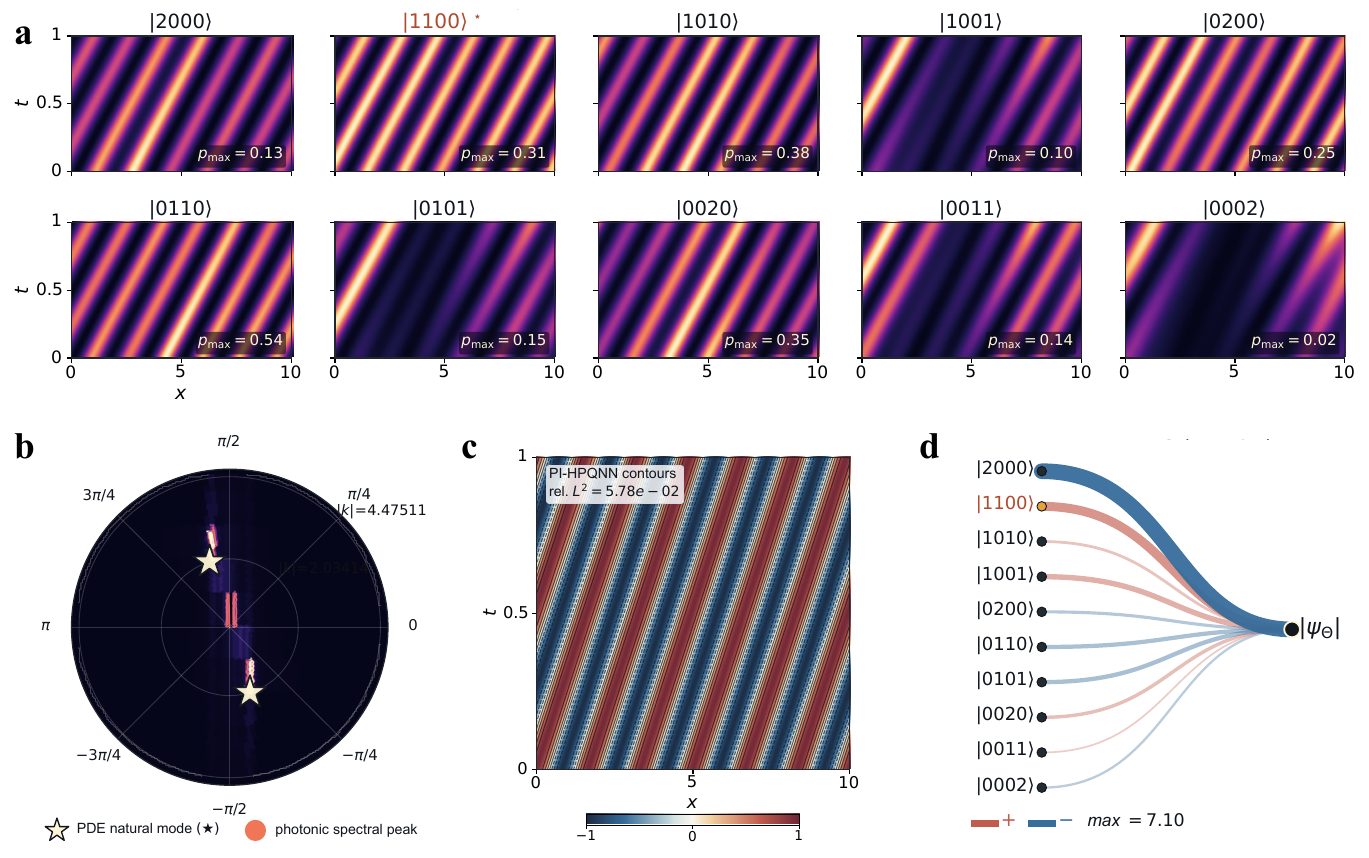}
\caption{Fock-space interference atlas for the high-complexity nonlinear Schr\"odinger benchmark at \(n=6\). \textbf{a}, Measured Fock probabilities \(p_\alpha(x,t)\) for the ten two-photon basis states \(\alpha\in\{|2000\rangle,\ldots,|0002\rangle\}\) of the trained PI-HPQNN layer. The highlighted state \(\alpha^\star=|1100\rangle\) is the physical input preparation channel; the full probability vector forms the measured feature representation supplied to the readout. \textbf{b}, Polar spectral fingerprint of the aggregate Fock-channel spectrum, with stars marking the natural PDE mode directions and a dominant photonic wavenumber \(|k|\approx 4.48\); the photonic spectral peaks concentrate near the PDE modes although no spectral labels are supplied during training. \textbf{c}, PI-HPQNN reconstruction of the real component \(u^\star=\Re\psi^\star\) of the NLS field overlaid on the reference (representative single-seed example with rel.~\(L^2 = 5.78\times 10^{-2}\)). \textbf{d}, Readout-coupling diagram from the ten measured Fock channels to the predicted amplitude \(|\psi_\Theta|\); line width encodes readout sensitivity, with positive (red) and negative (blue) contributions and a maximum coupling of \(7.10\), showing that only a structured subset of measured channels carries most of the sensitivity. Together, the panels show that the photonic layer supplies phase-aligned physical measurement features rather than an unconstrained latent code.}
\label{fig:fock_atlas_nls}
\end{figure}

We next inspected what the learned representation actually is. Figure~\ref{fig:fock_atlas_nls} opens the measured Fock-space features for the high-complexity NLS case, where the exact solution is a travelling complex phase field. The ten panels in Fig.~\ref{fig:fock_atlas_nls}a are not generic hidden activations. They are photon-number probabilities \(p_\alpha(x,t)\) produced by the trained phase encoder, multi-mode interferometer and Fock measurement in the two-photon sector. Each measurement channel forms a structured oscillatory field, and different Fock outcomes emphasize different phase bands. The intermediate variables are therefore normalized physical measurement probabilities with a defined optical origin.

Figure~\ref{fig:fock_atlas_nls}b--d shows how these measured probabilities become a residual-learning representation. The polar spectral fingerprint in Fig.~\ref{fig:fock_atlas_nls}b concentrates along the natural NLS mode directions, and the marked PDE modes fall on the same angular support as the dominant photonic peaks. No spectral labels are supplied during training; this alignment is produced by optimizing the PDE residual and constraints through the phase encoder, interferometer and readout. The reconstruction map in Fig.~\ref{fig:fock_atlas_nls}c closes the loop: the trained photonic representation produces contours that track the real component of the NLS field and preserve the travelling phase pattern. The readout-coupling diagram in Fig.~\ref{fig:fock_atlas_nls}d shows selective use of the measured channels, with a small number of Fock outcomes carrying most of the sensitivity.

This analysis supplies the first mechanistic pillar of the Results: the useful variables are measured photonic moments, not unconstrained latent units. Phase is encoded as optical phase, mixed by interference, exposed through Fock probabilities and selected by the readout to reconstruct a phase-consistent PDE solution. The quantities that become informative are precisely those controlled by photonic design choices, including input Fock state, interferometer topology, data reuploading depth and measurement map. Structure ablations over these choices show that accuracy changes when the accessible measured moments change.

\unnumlead{Training the interference drives the representation}

The Fock atlas shows what the trained representation looks like. The second mechanistic question is whether that representation depends on learned interference or only on access to a random Fock-probability interface. To test this, we used the NLS benchmark at \(n=4\), where the photonic layer is active but optimization remains stable enough for controlled ablations. Three variants were compared over five seeds: the full trained PI-HPQNN, a frozen-random photonic circuit with only the classical stem and readout trained, and a frozen-shuffled circuit obtained by permuting the trained photonic tensors while preserving their value distribution.

The result is decisive (Table~\ref{tab:frozen_photonic_controls}). Freezing a random circuit increases the coupled-field error from \(1.95\%\) to \(3.90\%\), and shuffling the trained circuit increases it to \(3.47\%\). The amplitude error follows the same pattern. The photonic layer is therefore not acting as a disposable random-feature generator, and the gain is not explained by parameter scale alone. The representation works because the interference pattern is learned by the residual.

\begin{table}[!htbp]
\centering
\resulttablefont
\caption{Frozen-circuit controls on the NLS benchmark at \(n=4\). Errors are relative \(L^2\) values reported as mean \(\pm\) sample standard deviation over five seeds. The frozen-random condition freezes a randomly initialized photonic layer and trains only the classical components. The frozen-shuffled condition freezes a trained photonic layer after randomly permuting each parameter tensor, preserving the parameter distribution while destroying learned structure.}
\label{tab:frozen_photonic_controls}
\setlength{\tabcolsep}{5pt}
\renewcommand{\arraystretch}{1.08}
\begin{tabular}{lcc}
\toprule
Condition & \((u,v)\) error & \(|\psi|\) error \\
\midrule
Trained PI-HPQNN & \(\best{0.0195 \pm 0.0032}\) & \(\best{0.0118 \pm 0.0018}\) \\
Frozen-random photonic layer & \(0.0390 \pm 0.0115\) & \(0.0241 \pm 0.0065\) \\
Frozen-shuffled photonic layer & \(0.0347 \pm 0.0054\) & \(0.0226 \pm 0.0037\) \\
\bottomrule
\end{tabular}
\end{table}

\unnumlead{Fock-probability measurement stabilizes the readout under compound noise}

The third mechanistic question is whether the measured probability representation has a distinctive noise response. A clean-trained PI-HPQNN was subjected at inference time to surrogate photonic perturbations representing transmission loss, reduced source indistinguishability and finite-shot sampling on the measured Fock-probability vector. The clean PI-HPQNN error on the same \(n=4\) NLS grid is \(0.0281\), compared with \(0.0353\) for the clean PINN baseline. Under single-axis perturbations, PI-HPQNN remains within \(2\times\) of its clean error for \(\eta\ge0.99\), \(V\ge0.97\) and \(N_{\mathrm{shots}}\ge10^5\).

The sharper comparison is against a fully retrained qubit-style HQNN evaluated under the same perturbation model. In the clean setting, this HQNN attains lower error than PI-HPQNN (\(0.0169\) versus \(0.0281\)). Under joint perturbations, however, the ordering reverses at every tested operating point (Table~\ref{tab:compound_noise_controls}). PI-HPQNN gives lower errors under mild, moderate and severe combined noise. The result does not claim hardware-level robustness, because the perturbations are injected at inference time into the measured feature representation. It shows the representation-level point needed here: normalized Fock probabilities can trade some clean accuracy for a more stable readout when transmission, indistinguishability and shot effects act together.

\begin{table}[!htbp]
\centering
\resulttablefont
\caption{Evaluation-time compound-noise stress test on the NLS benchmark at \(n=4\). Errors are relative \(L^2\) values for the coupled field \((u,v)\). Noise is injected at inference time into the measured feature representation; detailed single-axis sweeps are available with the source data.}
\label{tab:compound_noise_controls}
\setlength{\tabcolsep}{5pt}
\renewcommand{\arraystretch}{1.08}
\begin{tabular}{lccccc}
\toprule
Setting & \(\eta\) & \(V\) & \(N_{\mathrm{shots}}\) & HQNN & PI-HPQNN \\
\midrule
Ideal & 1.00 & 1.00 & \(\infty\) & \(\best{0.0169}\) & \(0.0281\) \\
Mild & 0.99 & 0.97 & \(10^5\) & \(0.0840\) & \(\best{0.0630}\) \\
Moderate & 0.95 & 0.95 & \(10^4\) & \(0.1858\) & \(\best{0.1266}\) \\
Severe & 0.90 & 0.90 & \(10^3\) & \(0.3794\) & \(\best{0.2741}\) \\
\bottomrule
\end{tabular}
\end{table}
\FloatBarrier

% ============================================================

\section{Discussion}
% ============================================================

This study identifies photonic quantum measurement as a trainable representation-learning principle for scientific machine learning. The central contribution is not that a quantum module is appended to a physics-informed neural network. It is that the measured photonic circuit defines the function space on which the PDE residual is minimized. Coordinates are encoded as optical phases, mixed by multi-photon interference, exposed through Fock-space measurement and optimized by the same residual, boundary, data and parameter-identification losses as the classical baselines. The learned object is therefore not only a neural approximation. It is a measured physical representation shaped by the governing equation.

The evidence supports this interpretation because the advantage appears selectively, not uniformly. PI-HPQNN is strongest in the high-frequency Poisson and Helmholtz problems, the most oscillatory forced Sine--Gordon case, the high-complexity nonlinear Schr\"odinger case and the rapidly phase-varying wave benchmark. In smoother regimes, classical coordinate and Fourier-feature baselines remain preferable. This phase-complexity transition is important because it turns a benchmark comparison into a design rule: photonic trial spaces are most useful when residual derivatives amplify phase mismatch. The inverse Burgers and Euler experiments add a complementary setting, showing that measured photonic features can help sparse observations constrain latent physical parameters, even when pointwise field interpolation is not always the winning metric.

The controls and diagnostics explain why this design rule is not simply a parameter-count effect. The Fock atlas shows that the intermediate variables are structured photon-number probabilities, not generic hidden activations. Frozen and shuffled circuit controls show that the performance gain depends on learned interference rather than random Fock features alone. The noise stress test adds a representation-level robustness observation: a qubit-style HQNN can be more accurate in the ideal setting, but Fock-probability readout is more stable under compound perturbations in the tested surrogate model. Together with the structure ablations, these results indicate that accuracy is controlled by which measured spectral moments the photonic layer exposes to the decoder. The broader lesson is that physical measurement can be part of the learned representation, provided that the measurement basis is aligned with the structure amplified by the loss.

\unnumlead{Relation to other physical computing substrates}
Photonic computing has been pursued along several distinct routes, and PI-HPQNN occupies a specific position among them. Analog photonic neural networks based on Mach--Zehnder interferometer meshes use photonics to accelerate matrix multiplication, with nonlinearity supplied electronically or through dedicated optical activation devices \cite{ashtiani2026integrated_pnn_bp}. Diffractive and \(4f\)-Fourier processors exploit free-space propagation as a fixed optical Fourier transform, providing a static spectral basis whose accessible frequencies do not adapt to the governing equation \cite{lin2018diffractive,wetzstein2020inference}. Photonic reservoir computers couple untrained nonlinear photonic dynamics to a trained linear readout, leaving the dynamical layer outside the optimization \cite{vandoorne2014reservoir,larger2017reservoir}.

PI-HPQNN differs from each of these routes in the role assigned to the optical system. Its computational primitive is multi-photon interference inside a fixed-photon-number subspace. Fock-space measurement converts coherent amplitudes into real nonlinear spectral moments, and these measured features are trainable through optical phase encoding, programmable interferometers, reuploading depth and input Fock state. The resulting object is neither a linear matrix multiplier, nor a fixed Fourier basis, nor an untrained dynamical reservoir. It is a trainable measured spectral function space. This distinction matters for scientific computing because oscillatory PDE residuals do not mainly require faster matrix products. They require a trial space whose derivative structure tracks the equation's phase content, and the feature-space theorem in Methods formalizes this trial space.

\unnumlead{Limitations}
Three limitations bound the present results. First, all experiments are performed in differentiable classical simulation of the photonic circuit. The reported wall-clock costs therefore reflect software simulation of multi-photon interference rather than native photonic execution, and the parameter-efficiency advantage cannot yet be converted into a hardware-throughput claim. The robustness study is an evaluation-time surrogate noise test on measured feature vectors. End-to-end training under hardware noise will require parameter-shift gradient rules, calibrated photonic sampling or density-matrix simulation, none of which is exercised here.

Second, the forward benchmarks use prescribed analytical references so that residual-based learners can be compared under controlled phase-complexity ladders. Broader testing on independent numerical references, stiff multi-scale physics and experimental observations remains future work. Third, the photonic representation is intrinsically quasi-Fourier. The feature-space and approximation theorems make the spectral nature of the trial space explicit, and PDEs whose solutions are not well represented by frequency-difference moments at the accessible photonic bandwidth are not expected to benefit from this architecture. These limitations sharpen rather than weaken the design rule. The photonic trial space should be considered when the governing residual amplifies phase error and the solution is broadly band-limited. Smooth low-frequency problems are already well served by classical coordinate networks, and the present benchmarks identify the crossover quantitatively for each equation family.

\unnumlead{Outlook}
The natural next step is native photonic hardware, where the same architecture must be tested under physical sampling, calibration drift, multi-photon source statistics, on-chip loss and finite detection efficiency. In parallel, the approximation picture should be sharpened. Useful targets include rules linking circuit depth, photon number, mode connectivity and measurement choice to a target Sobolev tail; convergence rates for the trainable nonlinear phase map; and bridges to neural-operator frameworks where a measured photonic feature space might serve as a discretization-invariant trial space. More broadly, the results suggest a general route for machine learning with trainable physical measurement processes, in which the representation is not only implemented on a device but learned through the device's native measurement. PDEs whose solutions are organized by phase provide a natural frontier for this idea: they can be approximated by representations generated, mixed and measured through phase itself.

% ============================================================

\section{Methods}
% ============================================================

The method is designed to isolate the role of the learned representation. All compared models use the same physics-informed objective, residual discretization, constraint terms, data losses and inverse-parameter losses. PI-HPQNN changes only the differentiable map from physical coordinates to the PDE state. This map lifts coordinates into trainable optical phases, evolves a fixed-photon-number photonic circuit, measures Fock-space probabilities and decodes the measured probabilities into the solution variables. We first define the shared objective, then specify the photonic architecture, training protocol, evaluation metrics and control experiments. We then give the spectral mechanism that links phase encoding, interference and measurement to residual-based PDE learning.

\methodlead{Physics-informed objective}

All models were trained as differentiable trial functions for the same physics-informed optimization problem. Let $Q$ denote the input domain, with $Q=\Omega$ for steady problems and $Q=\Omega\times(0,T]$ for time-dependent problems. The independent variable is $\boldsymbol{\xi}$, for example $\boldsymbol{\xi}=(x,y)^\top$ in two-dimensional elliptic problems and $\boldsymbol{\xi}=(x,t)^\top$ in one-dimensional time-dependent problems. A PDE with optional physical parameters $\boldsymbol{\lambda}$ is written as
\begin{equation}
    \mathcal{N}[u](\boldsymbol{\xi};\boldsymbol{\lambda})=0,
    \qquad \boldsymbol{\xi}\in Q,
    \label{eq:methods_pde}
\end{equation}
with boundary, initial or periodic constraints collected as
\begin{equation}
    \mathcal{C}[u](\boldsymbol{\xi})=c(\boldsymbol{\xi}),
    \qquad \boldsymbol{\xi}\in\Gamma .
    \label{eq:methods_constraint}
\end{equation}
The approximation $\widehat{u}_{\Theta}$ is substituted into the residual and all derivatives are evaluated by automatic differentiation.

The empirical objective used for PI-HPQNN, PINN, FPINN and HQNN is
\begin{equation}
    \mathcal{L}_{\mathrm{total}}
    =
    \omega_r\mathcal{L}_{r}
    +
    \omega_c\mathcal{L}_{c}
    +
    \omega_d\mathcal{L}_{d}
    +
    \omega_p\mathcal{L}_{p},
    \label{eq:compact_pinn_loss}
\end{equation}
where unused terms are omitted. The residual and constraint losses are
\begin{equation}
    \mathcal{L}_{r}
    =
    \frac{1}{N_r}\sum_{j=1}^{N_r}
    \left\|
    \mathcal{N}[\widehat{u}_{\Theta}](\boldsymbol{\xi}_j^r;\boldsymbol{\lambda})
    \right\|_2^2,
    \qquad
    \mathcal{L}_{c}
    =
    \frac{1}{N_c}\sum_{j=1}^{N_c}
    \left\|
    \mathcal{C}[\widehat{u}_{\Theta}](\boldsymbol{\xi}_j^c)-c(\boldsymbol{\xi}_j^c)
    \right\|_2^2 .
    \label{eq:res_constraint_loss}
\end{equation}
The data term averages $\|\widehat{u}_{\Theta}(\boldsymbol{\xi}_j^d)-u_j^{\mathrm{obs}}\|_2^2$ over observations. The norm is the squared Euclidean norm at each point; for vector-valued residuals and for the coupled real representation of complex-valued fields, it is applied to the component vector. In inverse problems, selected entries of $\boldsymbol{\lambda}$ are included in the trainable variables; positivity-constrained coefficients are optimized through an unconstrained parameterization, such as $\nu=\exp(\widetilde{\nu})$ for the inverse Burgers viscosity. The loss weights, sampling sets and optimization schedules are kept identical across models within each benchmark, so the comparisons isolate the learned representation rather than the physics-informed objective.

The same residual implementation is used for all models. PI-HPQNN changes only the differentiable map $\boldsymbol{\xi}\mapsto\widehat{u}_{\Theta}(\boldsymbol{\xi})$ that is substituted into Eqs.~\eqref{eq:methods_pde}--\eqref{eq:methods_constraint}; it does not change the collocation discretization, boundary or initial loss, data loss, or inverse-parameter objective. Higher-order derivatives in elliptic, wave and dispersive equations are obtained by repeated automatic differentiation of the complete trial function with respect to the physical coordinates.

\methodlead{Photonic quantum neural field architecture}

PI-HPQNN replaces the ordinary coordinate-network representation by a classical--photonic--classical composition. Given an input coordinate $\boldsymbol{\xi}$, a classical stem lifts it to a hidden representation, a trainable encoder converts that representation into wrapped optical phases, a multi-mode photonic circuit evolves a fixed Fock input state, Fock-basis probabilities are measured and a classical readout maps those probabilities to the PDE state. The trial function is
\begin{equation}
\begin{aligned}
    \widehat{u}_{\Theta}(\boldsymbol{\xi})
    & =
    R_{\theta_r}
    \circ
    \mathcal{M}
    \circ
    U_{\theta_q}
    \circ
    E_{\theta_e}
    \circ
    S_{\theta_s}(\boldsymbol{\xi}),
\end{aligned}
    \label{eq:hpqnn_composition}
\end{equation}
where $S_{\theta_s}$ is the classical lifting stem, $E_{\theta_e}$ is the trainable phase encoder, $U_{\theta_q}$ is the MerLin-based photonic layer, $\mathcal{M}$ is Fock-probability readout and $R_{\theta_r}$ is the classical decoder.

The stem is a one-layer tanh network,
\begin{equation}
    \mathbf{h}(\boldsymbol{\xi})
    =
    \tanh(\mathbf{W}_s\boldsymbol{\xi}+\mathbf{b}_s)
    \in\mathbb{R}^{m_s},
    \qquad m_s=20 .
    \label{eq:stem_map}
\end{equation}
The hidden vector is projected to $D_q=2dL$ phase variables, with $d=4$ encoded angles and $L=2$ data-reuploading blocks in the default setting. The phase map is
\begin{equation}
    \boldsymbol{\varphi}(\boldsymbol{\xi})
    =
    \operatorname{wrap}_{[-\pi,\pi]}
    \left[
    \boldsymbol{\gamma}\odot
    (\mathbf{W}_e\mathbf{h}(\boldsymbol{\xi})+\mathbf{b}_e)
    +
    \boldsymbol{\beta}
    \right],
    \label{eq:angle_wrap}
\end{equation}
where $\boldsymbol{\gamma}$ and $\boldsymbol{\beta}$ are trainable gains and phase offsets. The wrapping operation is implemented as $\operatorname{atan2}(\sin a,\cos a)$ componentwise, which maps the encoded phases to the principal interval while remaining differentiable almost everywhere. The spectral expressions below depend on $e^{i\varphi}$; since $e^{i\operatorname{wrap}(a)}=e^{ia}$, the phase branch does not change the trigonometric factors, and the branch points are a measure-zero issue for automatic differentiation.

The encoded phase vector is partitioned according to the reuploading structure,
\begin{equation}
    \boldsymbol{\varphi}
    =
    \left(
    \boldsymbol{\varphi}_{1}^{(1)},
    \boldsymbol{\varphi}_{1}^{(2)},
    \ldots,
    \boldsymbol{\varphi}_{L}^{(1)},
    \boldsymbol{\varphi}_{L}^{(2)}
    \right),
    \qquad
    \boldsymbol{\varphi}_{\ell}^{(r)}\in\mathbb{R}^{d}.
    \label{eq:phase_partition}
\end{equation}
For the default $L=2$ and $d=4$, this gives four four-dimensional phase groups. Reuploading inserts input-dependent phases at more than one depth of the circuit, allowing trainable optical transformations and coordinate-dependent phase shifts to alternate during the forward pass.

In the spectral statements below, the relevant count is the number of input-dependent encoding stages rather than the number of reuploading blocks alone. We write
\begin{equation}
    L_{\mathrm{enc}}=qL,
    \qquad q=2
    \label{eq:encoding_stage_count}
\end{equation}
for the implementation used here, because each block contains two phase-encoding stages.

\methodlead{Photonic circuit, measurement and readout}

The default photonic layer is a four-mode, two-photon discrete-variable circuit. With total photon number $n$ and $M$ modes, the fixed-photon-number Fock subspace has size
\begin{equation}
    R(M,n)=\binom{M+n-1}{n}.
    \label{eq:fock_dimension}
\end{equation}
The main experiments use $M=4$, $n=2$ and input state $|\mathbf{n}_0\rangle=|1,1,0,0\rangle$, giving $R(4,2)=10$ Fock probabilities. Each reuploading block contains trainable single-mode rotations, two phase-encoding stages, two supermode interference layers and two trainable two-mode coupling layers. The phase groups in Eq.~\eqref{eq:phase_partition} are injected into the two encoding stages of each block. The standard layered topology is used for the main experiments; cross-mesh, cascade and alternate connectivities are used only in the topology ablation.

For an input $\boldsymbol{\xi}$, the photonic circuit prepares
\begin{equation}
    |\psi_{\Theta}(\boldsymbol{\xi})\rangle
    =
    U_{\theta_q}(\boldsymbol{\varphi}(\boldsymbol{\xi}))
    |\mathbf{n}_0\rangle .
    \label{eq:pqnn_state}
\end{equation}
The measured features are Fock-basis probabilities
\begin{equation}
    p_k(\boldsymbol{\xi})
    =
    \left|
    \langle \mathbf{n}^{(k)}|\psi_{\Theta}(\boldsymbol{\xi})\rangle
    \right|^2,
    \qquad k=1,\ldots,K,
    \label{eq:fock_probability}
\end{equation}
with $K=10$ for the default full probability vector. The full probability vector is bounded and normalized, $p_k\ge 0$ and $\sum_k p_k=1$, before it enters the decoder. The decoder is a one-hidden-layer tanh network,
\begin{equation}
    \widehat{u}_{\Theta}(\boldsymbol{\xi})
    =
    \mathbf{W}_{r,2}
    \tanh(\mathbf{W}_{r,1}\mathbf{p}_{\Theta}(\boldsymbol{\xi})+\mathbf{b}_{r,1})
    +
    \mathbf{b}_{r,2},
    \qquad m_r=20 ,
    \label{eq:readout_network}
\end{equation}
where the final affine layer has one output for scalar fields and multiple coupled real outputs for vector or complex-valued systems. For scalar two-coordinate benchmarks, the default parameter count is $60$ stem parameters, $368$ encoder parameters, $96$ photonic-layer parameters and $241$ readout parameters, for a total of $765$ trainable parameters. Vector-valued outputs and inverse coefficients add only the corresponding final-output or physical-parameter variables, giving the nearby parameter counts reported in Results.

All simulations are differentiable through the classical photonic simulator. Gradients with respect to the classical parameters and photonic parameters are obtained through the automatic-differentiation graph; parameter-shift rules are not used in the reported simulations. For first derivatives, the residual differentiates through the composition
\begin{equation}
    \nabla_{\boldsymbol{\xi}}\widehat{u}_{\Theta}
    =
    J_R J_{\mathcal{M}} J_U J_E J_S ,
    \label{eq:trial_chain_rule}
\end{equation}
with Jacobians evaluated along the forward pass in Eq.~\eqref{eq:hpqnn_composition}. The trainable variables are
\begin{equation}
    \Theta=
    \{\theta_s,\mathbf{W}_e,\mathbf{b}_e,\boldsymbol{\gamma},\boldsymbol{\beta},
    \theta_q,\theta_r,\boldsymbol{\lambda}\},
    \label{eq:trainable_set}
\end{equation}
where $\boldsymbol{\lambda}$ appears only for inverse problems.

\methodlead{Training, benchmarks and evaluation}

Training uses Adam to reach a stable basin followed by L-BFGS residual refinement, following common PINN practice. Within each benchmark, PI-HPQNN and all baselines use the same collocation points, boundary or initial points, observation sets, loss weights and optimizer budgets. The baseline models are a standard tanh PINN, a Fourier-feature PINN (FPINN) with fixed coordinate features, and a hybrid quantum neural network (HQNN) baseline. The reported trainable-parameter counts include all learned neural, quantum, photonic and inverse physical parameters. These choices make the comparison a test of the coordinate-to-field representation rather than a test of different residual objectives or data budgets.

Benchmark-specific domains, exact solutions, forcing terms, sampling sizes, loss weights, training schedules and repeated-run conventions are summarized in Results and specified here. The forward benchmark suite includes Poisson, wave, Helmholtz, forced Sine--Gordon and nonlinear Schr\"odinger equations. These benchmark equations are used here as controlled phase-complexity tests rather than as replacements for specialized solvers; for nonlinear Schr\"odinger/Gross--Pitaevskii and related highly oscillatory wave models, time-splitting, spectral and multiscale integrators provide well-established numerical foundations \cite{bao2003nls_tss,bao2003gpe_jcp,bao2013bec_review,bao2014kg_mti}. The inverse suite includes Burgers viscosity recovery and Euler coefficient recovery from sparse observations, with additional $5\%$ noise settings where specified. Noisy observations are generated by adding Gaussian perturbations scaled to the observation magnitude used in the corresponding benchmark.

For complex-valued equations, the real and imaginary parts are represented as coupled real outputs. The main accuracy metric is the grid relative $L^2$ error,
\begin{equation}
    \varepsilon_{L^2}
    =
    \frac{\left(\sum_i \|\widehat{u}_{\Theta}(\boldsymbol{\xi}_i)-u^\star(\boldsymbol{\xi}_i)\|_2^2\right)^{1/2}}
    {\left(\sum_i \|u^\star(\boldsymbol{\xi}_i)\|_2^2\right)^{1/2}},
    \label{eq:relative_l2_metric}
\end{equation}
computed on the evaluation grid used for each benchmark. For NLS, both the coupled real--imaginary error and the amplitude error are reported. For inverse problems, state errors are reported together with relative errors in the identified coefficients. Repeated-run statistics are reported as mean $\pm$ sample standard deviation when repeated runs are available.

Wall-clock timings are classical simulation times measured for the recorded implementations. They are reported to document computational cost under differentiable simulation and should not be interpreted as native photonic hardware timings. Parameter counts are computed from trainable weights only and include learned physical coefficients in inverse problems.

\methodlead{Representation controls and feature-level perturbations}

Frozen-circuit controls use the NLS benchmark at \(n=4\) and keep the classical training protocol unchanged. In the trained condition, the phase encoder, photonic circuit and readout are optimized jointly. In the frozen-random condition, the photonic circuit is initialized randomly and then held fixed while the classical stem, encoder and readout remain trainable. In the frozen-shuffled condition, the trained photonic parameter tensors are randomly permuted tensor by tensor, preserving their marginal value distributions while destroying learned optical structure; the shuffled photonic circuit is then held fixed. These controls test whether performance depends on learned interference rather than on access to a ten-dimensional Fock-probability interface alone.

The evaluation-time noise study perturbs the measured feature vector after clean training. Let \(\mathbf{p}(\boldsymbol{\xi})\in\mathbb{R}^{K}\) be the ideal Fock-probability vector and \(\mathbf{u}_K=K^{-1}\mathbf{1}\) the uniform distribution. Transmission loss and reduced indistinguishability are represented by the feature-level contraction
\begin{equation}
    \mathbf{p}_{\eta,V}(\boldsymbol{\xi})
    =
    V\left[
    \eta \mathbf{p}(\boldsymbol{\xi})+(1-\eta)\mathbf{u}_K
    \right]
    +(1-V)\mathbf{u}_K ,
    \label{eq:methods_surrogate_noise_probability}
\end{equation}
where \(\eta=1\) and \(V=1\) recover the clean feature vector. Finite-shot measurement is represented by multinomial sampling,
\begin{equation}
    \mathbf{c}(\boldsymbol{\xi})
    \sim
    \operatorname{Multinomial}
    \left(
    N_{\mathrm{shots}},
    \mathbf{p}_{\eta,V}(\boldsymbol{\xi})
    \right),
    \qquad
    \widehat{\mathbf{p}}_{\eta,V,N}
    =
    \frac{\mathbf{c}}{N_{\mathrm{shots}}},
    \label{eq:methods_surrogate_shot_noise}
\end{equation}
with \(N_{\mathrm{shots}}=\infty\) defined as the exact vector \(\mathbf{p}_{\eta,V}\). The noisy feature vector is passed through the trained readout without updating model parameters. The same feature-level convention is applied to the HQNN baseline by mapping each expectation feature \(z_j\in[-1,1]\) to a Bernoulli probability \(q_j=(1+z_j)/2\), applying the same contraction and binomial sampling, and mapping back by \(\widehat{z}_j=2\widehat{q}_j-1\). This is a surrogate representation-level stress test; it does not simulate coherent loss modes, detector response, multiphoton source statistics or noisy-gradient training.

\methodlead{Photonic spectral mechanism}

The photonic layer defines a measured trigonometric representation rather than an opaque feature extractor. Expanding the output state in the fixed-photon-number Fock basis gives
\begin{equation}
    |\psi_{\Theta}(\boldsymbol{\xi})\rangle
    =
    \sum_{\boldsymbol{\alpha}\in\mathcal{F}_{M,n}}
    A_{\boldsymbol{\alpha}}(\boldsymbol{\xi};\theta_q)
    |\boldsymbol{\alpha}\rangle ,
    \qquad
    A_{\boldsymbol{\alpha}}
    =
    \sum_{\boldsymbol{\omega}\in\Omega_{\boldsymbol{\alpha}}}
    c_{\boldsymbol{\alpha},\boldsymbol{\omega}}(\theta_q)
    e^{i\boldsymbol{\omega}\cdot\boldsymbol{\varphi}(\boldsymbol{\xi})}.
    \label{eq:methods_amplitude_expansion}
\end{equation}
Here $\mathcal{F}_{M,n}$ is the $n$-photon Fock basis over $M$ modes, and the accessible frequency index set $\Omega_{\boldsymbol{\alpha}}$ is determined by the input Fock state, photon number, circuit topology, phase-encoding locations and reuploading depth. Trainable interferometers set the coefficients $c_{\boldsymbol{\alpha},\boldsymbol{\omega}}(\theta_q)$, while the classical encoder sets the phase map $\boldsymbol{\varphi}(\boldsymbol{\xi})$ and therefore the effective local frequencies.

Fock probability measurement turns coherent optical amplitudes into real nonlinear spectral moments. Since $p_{\boldsymbol{\alpha}}=|A_{\boldsymbol{\alpha}}|^2$, Eq.~\eqref{eq:methods_amplitude_expansion} yields
\begin{equation}
    p_{\boldsymbol{\alpha}}(\boldsymbol{\xi})
    =
    \sum_{\boldsymbol{\omega},\boldsymbol{\omega}'\in\Omega_{\boldsymbol{\alpha}}}
    c_{\boldsymbol{\alpha},\boldsymbol{\omega}}
    \overline{c_{\boldsymbol{\alpha},\boldsymbol{\omega}'}}
    e^{i(\boldsymbol{\omega}-\boldsymbol{\omega}')
    \cdot
    \boldsymbol{\varphi}(\boldsymbol{\xi})}.
    \label{eq:methods_measurement_moments}
\end{equation}
Thus the measured coordinates supplied to the decoder are frequency-difference moments created by multi-photon interference. Since probabilities are real, the same statement may be read equivalently as a real span of the corresponding sine and cosine moments. This gives a compact feature-space theorem for the photonic representation.

\begin{theorem}[Photonic feature-space theorem]
\label{thm:methods_feature_space}
\leavevmode\par\nobreak
Consider an $L$-block photonic circuit with $M$ modes and $n$ photons, using angle encoding $\boldsymbol{\varphi}(\boldsymbol{\xi})$ and $L_{\mathrm{enc}}=qL$ input-dependent encoding stages. The trainable optical mixing layers are input-independent photon-number-preserving unitaries. Each Fock-probability feature $p_{\boldsymbol{\alpha}}(\boldsymbol{\xi})$ lies in the finite trigonometric span
\begin{equation}
    p_{\boldsymbol{\alpha}}
    \in
    \operatorname{span}
    \left\{
    e^{i(\boldsymbol{\omega}-\boldsymbol{\omega}')
    \cdot
    \boldsymbol{\varphi}(\boldsymbol{\xi})}
    :
    |\boldsymbol{\omega}|_1\le L_{\mathrm{enc}}n,\ 
    |\boldsymbol{\omega}'|_1\le L_{\mathrm{enc}}n
    \right\},
    \label{eq:methods_feature_span}
\end{equation}
with the actually expressed subset controlled by the input Fock state, photonic mesh, encoding locations and measurement choice.
\end{theorem}

Theorem~\ref{thm:methods_feature_space} is the first link in the mathematical story. Phase encoding supplies elementary Fourier factors, multi-photon interference mixes these factors in optical amplitudes, Fock measurement forms nonlinear frequency-difference moments and the decoder uses those measured moments as the representation optimized by the physics-informed residual. Photon number and the number of encoding stages control the ambient bandwidth, while the trainable interferometers determine which spectral moments are expressed.

This feature span is naturally matched to residual-based PDE learning. For a measured phase feature $\chi_{\mathbf{m}}(\boldsymbol{\xi})=e^{i\mathbf{m}\cdot\boldsymbol{\varphi}(\boldsymbol{\xi})}$,
\begin{equation}
    \nabla_{\boldsymbol{\xi}}\chi_{\mathbf{m}}
    =
    i\left[J_{\boldsymbol{\varphi}}(\boldsymbol{\xi})^T\mathbf{m}\right]
    \chi_{\mathbf{m}}(\boldsymbol{\xi}),
    \label{eq:methods_phase_derivative}
\end{equation}
so differentiation keeps the feature inside the same oscillatory function class while scaling it by the learned local wave vector. Locally, when $\boldsymbol{\varphi}(\boldsymbol{\xi})\simeq A\boldsymbol{\xi}+\mathbf{c}$, the effective physical frequency is $A^T\mathbf{m}$. The PDE residual can therefore optimize not only decoder weights, but also the encoded frequency scale, phase offsets and coherent mode mixing that generate the measured moments.

The same structure gives a compact approximation statement for the measured photonic representation.

\begin{theorem}[Photonic spectral approximation theorem]
\label{thm:methods_spectral_approximation}
\leavevmode\par\nobreak
Let \(u\in H^s(\mathbb{T}^{d_\xi})\), \(s>d_\xi/2\). Consider the idealized affine phase map \(\boldsymbol{\varphi}(\boldsymbol{\xi})=A\boldsymbol{\xi}+\mathbf{c}\) associated with a PI-HPQNN feature map, and let \(P_K\) denote the \(L^2\)-orthogonal projection onto
\(\mathcal{T}_K=\operatorname{span}\{e^{i\mathbf{k}\cdot\boldsymbol{\xi}}:\|\mathbf{k}\|_{\infty}\le K\}\),
with \(K\simeq\kappa L_{\mathrm{enc}}n\), where \(\kappa\) absorbs the physical-frequency scale of the affine phase map and the constant factor introduced by frequency-difference measurement. Let \(\mathcal{V}_{L,n}\) be the measured photonic feature span followed by a linear readout, and define the truncated-mode projection defect
\begin{equation}
    \delta_K(u;\mathcal{V}_{L,n})
    =
    \inf_{\widehat{v}\in\mathcal{V}_{L,n}}
    \|P_Ku-\widehat{v}\|_{L^2}.
    \label{eq:methods_projection_defect}
\end{equation}
Then
\begin{equation}
    \inf_{\widehat{v}\in\mathcal{V}_{L,n}}
    \|u-\widehat{v}\|_{L^2}
    \le
    C_{s,d_\xi}\|u\|_{H^s}(\kappa L_{\mathrm{enc}}n)^{-s}
    +
    \delta_K(u;\mathcal{V}_{L,n}),
    \label{eq:methods_fourier_tail}
\end{equation}
\end{theorem}

Theorem~\ref{thm:methods_spectral_approximation} makes the second link explicit without assuming that a finite set of measured probabilities spans every Fourier mode independently. The first term is the usual Sobolev Fourier tail; the second term measures whether the photonic probabilities exposed by the chosen phase map, optical mesh and measurement cover the truncated modes that matter for the target field. When this projection defect is small, the measured photonic representation inherits the spectral approximation behavior relevant to the PDE. The actual PI-HPQNN uses a nonlinear trainable phase map; locally, Eq.~\eqref{eq:methods_phase_derivative} replaces the affine matrix \(A\) by the learned Jacobian \(J_{\boldsymbol{\varphi}}(\boldsymbol{\xi})\), giving an adaptive version of the same spectral mechanism. The numerical ablations then test the third link of the story: measurement choice, topology, Fock input and reuploading depth change which moments are available, and these changes control accuracy in phase-sensitive PDE benchmarks. The nonlinear decoder used in the experiments provides additional trainable recombination beyond the linear span isolated in the theorem.

\section*{Data availability}

All benchmark data used in this study are generated from the analytical equations, domains, boundary or initial conditions and observation protocols described in the Methods. Source data for the figures, tables and additional analyses are available in the project repository at \url{https://github.com/linghujiale/PQNN}.

\section*{Code availability}

The code used to generate the benchmarks, train the PI-HPQNN and baseline models, run the representation controls and produce the figures is available at \url{https://github.com/linghujiale/PQNN}.

\section*{Acknowledgements}

Part of this work was carried out while J.L. was visiting the Department of Mathematics at the National University of Singapore. The authors are grateful to Weizhu Bao for insightful suggestions and valuable discussions, which helped sharpen the benchmark design and the interpretation of the numerical results.

\section*{Author contributions}

Y.W. conceived and supervised the study. J.L. and H.D. developed the implementation, performed the numerical experiments and prepared the figures. All authors analysed the results, contributed to the interpretation of the findings, wrote or revised the manuscript and approved the final version.

\section*{Competing interests}

The authors declare no competing interests.

\bibliography{sn-bibliography}

\end{document}